\documentclass[journal,draftcls,onecolumn,12pt,twoside]{IEEEtranTCOM}
\pdfoutput=1

\usepackage{amsmath}
\usepackage{amsfonts}
\usepackage{graphics, graphicx}
\usepackage{cite}
\usepackage{multirow}
\usepackage{url}
\usepackage{enumerate}
\usepackage{etoolbox}
\usepackage[ruled,norelsize]{algorithm2e}
\usepackage{numprint}
\usepackage{titlesec}
\usepackage{todonotes}
\usepackage{relsize}
\let\labelindent\relax
\usepackage{enumitem}
\usepackage{hyperref}
\usepackage{algorithmic}
\usepackage{titlesec}

\titlespacing*{\subsection}
{0pt}{0.5ex plus 1ex minus .2ex}{1ex plus .2ex}

\DeclareMathOperator*{\argmin}{arg\,min}

\newtheorem{conj}{Proposition}

\ifCLASSOPTIONcompsoc
\usepackage[caption=false,font=normalsize,labelfont=sf,textfont=sf]{subfig}
\else
\usepackage[caption=false,font=footnotesize]{subfig}
\fi

\newcommand{\norm}[1]{\left\lVert#1\right\rVert}

\newtheorem{theorem}{Theorem}

\begin{document}

\title{EQ-Net: Joint Deep Learning-Based Log-Likelihood Ratio Estimation and Quantization}
\author{Marius~Arvinte,~Ahmed~H.~Tewfik,~\IEEEmembership{Fellow,~IEEE,}~and~\\Sriram~Vishwanath,~\IEEEmembership{Senior~Member,~IEEE}\thanks{The authors are with the Department of Electrical and Computer Engineering, The University of Texas at Austin, Austin, TX 78712, USA.}}
\date{April 2021}

\maketitle

\vspace{-20mm}

\begin{abstract}
In this work, we introduce EQ-Net: a data-driven framework that solves both tasks of log-likelihood ratio (LLR) estimation and quantization. We motivate our approach with a theoretical proof about the dimensionality of the zero-forcing with successive interference (ZF-SIC) algorithm, showing that its estimate can be losslessly compressed. We then conjecture that this compression is optimal and propose a two-stage algorithm that uses supervised LLR compression as a pretext task for estimation. Our implementation is focused on low-latency, high-performance computational blocks achieved via deep neural networks. We carry out extensive experimental evaluation and demonstrate that our single architecture achieves state-of-the-art results on \textit{both} tasks when compared to previous methods, with gains in quantization efficiency as high as $20\%$ and reduced estimation latency by up to $60\%$ when measured on general purpose and graphical processing units (GPU). In particular, our approach reduces the GPU inference latency by more than two times in several multiple-input multiple-output (MIMO) configurations. Finally, we demonstrate that our scheme is robust to distributional shifts and retains a significant part of its performance when evaluated on 5G channel models that are never seen during training, as well as channel estimation errors.
\end{abstract}

\section{Introduction}
Digital multiple-input multiple-output (MIMO) communication systems operate by communicating a vector of discrete symbols across a single channel use. Two core tasks for the receiver in such a system consist in log-likelihood ratio (LLR) estimation (also commonly referred to as soft-output MIMO detection) and quantization. LLR estimation is a challenging practical problem due to the high complexity of the optimal solution and stringent latency requirements in 5G communication systems \cite{bennis2018ultrareliable}. For example, given a data frame with a duration of $1$ ms, each with $1000$ data sub-carriers, a base station could be faced with estimating LLR values from as many as tens of thousands of channel uses per data frame \cite{nr214}, each of these involving a costly search procedure. Solutions that are high-performance and low-latency are still an open problem and represent an active area of research. At the same time, near-optimal estimation algorithms are a central part of end-to-end performance in coded systems (e.g., low-density parity-check (LDPC) codes) \cite{albreem2019massive}. Log-likelihood ratio quantization is important in systems where energy and memory efficiency represent practical limitations. For example, in distributed communications systems \cite{kairouz2019advances}, low-resolution LLR quantization is required whenever relaying or feedback is involved, since system capacity represents a bottleneck. Quantization is also required in hybrid automatic repeat request (HARQ) schemes, where it is beneficial for the receiver to store LLR values from a failed transmission and to use a soft combining scheme \cite{frenger2001performance} to boost performance, making storage a bottleneck.

To address these issues, deep learning methods have emerged as promising candidates for aiding or completely replacing signal processing blocks in MIMO communication systems \cite{zhang2019deep,erpek2020deep}. These algorithms have an inherent computational advantage due to their parallel nature during inference and specialized hardware modules that have been developed. However, the robustness of such methods is still an open problem in the broader machine learning field \cite{akhtar2018threat} and has been recognized as an issue in digital communications as well \cite{kokalj2019adversarial}. In this paper, we introduce EQ-Net, a data-driven architecture that aims to solve the challenges of low-latency quantization and estimation, and robustness to distributional shifts, while still retaining superior end-to-end system performance.

\subsection{Related Work}
\subsubsection{LLR Estimation}
There are two variants of MIMO LLR estimation algorithms: hard- and soft-output. In this paper, we consider soft-output estimation algorithms, given that soft channel decoding is ubiquitous in practice. The V-BLAST algorithm \cite{wolniansky1998v} first introduces the idea of sequential estimation, and subsequent work has led to the core idea of the zero-forcing with successive interference cancellation (ZF-SIC) \cite{shen2004performance} as an algorithm for efficient detection. In this method, the system is reduced to its upper triangular form and data symbols are detected in a fixed order. Once a symbol is detected, it is assumed to be correct and subtracted from the remaining data streams. This leads to a low-complexity, but also low-performance method.

Sphere decoding \cite{hassibi2005sphere,studer2008soft} formulates the hard detection problem as a tree search algorithm and performs a greedy search. The algorithm supports either a hard \cite{guo2006algorithm} or a soft-output version \cite{studer2008soft}, where multiple candidate solutions are used to estimate likelihoods. The main drawback of algorithms in this family is that their end-to-end latency is prohibitively large even for moderately-sized systems due do their sequential nature. There also exists work that proposes parallelized versions of sphere decoding \cite{qi2010parallel}, as well as recent work that uses machine learning for initial search radius prediction \cite{mohammadkarimi2019deep}.

More recently, there has been work in model-based approaches for MIMO detection, where optimization steps (e.g., application of the gradient descent algorithm) are combined with deep learning. The work in \cite{he2019model} proposes OAMP-Net2 as a data-based extension to the OAMP detection algorithm \cite{ma2017orthogonal}, where the step sizes are treated as learnable parameters. This method has the advantage of a very small number of learnable parameters, but still suffers from increased end-to-end latency. The work in \cite{sholev2020neural} takes a similar approach, but replaces the fixed computations of the OAMP algorithm with fully learnable transforms (i.e., layers of a deep neural network), resulting in state-of-the-art results for MIMO LLR estimation. Similarly, the authors in \cite{shlezinger2020viterbinet} propose an architecture suitable for integration in the Viterbi decoding algorithm that also blends learnable transforms with classical algorithms. 

Finally, a different type of approach is the work in \cite{shental2019machine}, where a two-layer neural network learns a piecewise linear approximation of the log-likelihood ratios in single-input single-output (SISO) channels. However, it is unclear if this approach is easily extendable to MIMO scenarios, due to the increasing complexity of the problem.

\subsubsection{LLR Quantization}
The work in \cite{winkelbauer2015quantization} introduces an information-theoretic optimal data-based approach for quantizing log-likelihood values that are drawn from the same distribution (e.g., corresponding to bits found at the same position in Gray-coded digital quadrature amplitude modulation). The approach also has the advantage that it does not make any assumptions about the underlying channel model and an open-source data-based approach is offered for estimating optimal quantization levels in arbitrary channels. The work in \cite{zeitler2012quantize} proposes a solution for LLR quantization in relay systems based on maximizing the mutual information between two transmitters and one receiver.

The deep learning approach in \cite{arvinte2019deep} introduces a data-driven approach that trains an autoencoder with a carefully-chosen latent space dimension. The method leverages the redundancy between LLR values corresponding to a single channel use and achieves state-of-the-art quantization results for scalar interference channels. However, there remains the issue of extending this method to general MIMO scenarios, which is the major distinction between this paper and \cite{arvinte2019deep} in terms of quantization methods. Furthermore, this prior work does not consider the estimation problem at all.

\subsection{Contributions}
In this work, we introduce EQ-Net --- a deep learning framework that jointly tackles both log-likelihood ratio estimation (E) and quantization (Q) using a shared feature space. EQ-Net is the result of a data-driven methodology that learns a low-dimensional representation of the optimal solution and retains high estimation performance, organized as a two-stage training protocol. The practical implementation of our algorithm is achieved with a low-latency, non-recurrent deep learning architecture that uses compression as a \textit{pretext} task for the supervised learning of an estimator function. We release an open-source implementation of EQ-Net and all other considered methods\footnote{\url{https://github.com/mariusarvinte/eq-net}}.

Our main contributions are:
\begin{enumerate}[wide, labelindent=0pt]
    \item We prove that the ZF-SIC estimator implicitly outputs a compressed solution, i.e., it can be represented as a surjective function that maps a lower-dimensional vector to the estimated LLR vector in arbitrary MIMO channels. Based on this, we conjecture that this represents the optimal compression ratio for certain modulation orders, and provide experimental evidence to support this.
    \item We introduce a methodology for supervised training and evaluation a joint LLR estimation and quantization algorithm in MIMO scenarios. The approach is fully data-driven and involves a two-stage supervised training procedure: the first stage trains a quantization autoencoder (composed of an encoder and decoder), while the second stage only trains an estimation encoder and re-uses the quantization decoder. We perform ablation experiments to show that the two-stage training algorithm is essential for converging and that our approach is superior to single-stage supervised training algorithms for MIMO detection.
    \item We experimentally evaluate the end-to-end performance and end-to-end latency of EQ-Net in both tasks and show that it achieves state-of-the-art results. We compare our method with classical signal processing algorithms, as well as deep learning-based approaches. For quantization, we show gains of up to $20\%$ in compression efficiency and for estimation gains of up to $1$ dB in end-to-end performance, in several coded orthogonal frequency-division multiplexing (OFDM) MIMO scenarios. We demonstrate substantial latency improvements on both general and graphical processing units.
    \item We evaluate the robustness of EQ-Net against two different types of distributional shifts: inaccurate channel state information (CSI) and different train-test distributions of the channel matrix. In both cases, we show that EQ-Net achieves competitive performance compared to prior work based on deep learning.
\end{enumerate}

The remainder of the paper is structured as follows: Section \ref{sec:model} describes the system model and all assumptions we make. Section \ref{sec:theory} formulates and proves the theorem concerning the ZF-SIC solution and presents our conjecture regarding its optimality. Section \ref{sec:eqnet} motivates and introduces EQ-Net. Section \ref{sec:results} first provides experimental evidence for the proposed conjecture and validates our design choices. Then, we present experimental results on both quantization and estimation in two MIMO scenarios. Finally, Section \ref{sec:conc} discusses limitations and possible extensions of EQ-Net, and concludes the paper. We use lower- and uppercase bold characters to denote complex vectors and matrices, respectively. We use the symbol $\circ$ to denote function composition, i.e., $f \circ g = f(g(\cdot))$. We use the notation $\mathbf{X}^\mathrm{H}$ to denote the conjugate transpose of $\mathbf{X}$.

\section{System Model and Preliminaries}
\label{sec:model}
Assume a narrowband, instantaneous digital communication model given by \eqref{eq:system}. This encompasses several of the most practical scenarios, such as single carrier communication or an individual subcarrier of a MIMO-OFDM transmission, and is flexible enough to model various distributions of the MIMO channel matrix $\mathbf{H}$:
\begin{equation}
\label{eq:system}
    \mathbf{y} = \mathbf{Hx} + \mathbf{n},
\end{equation}

\noindent where $\mathbf{x} \in \mathbb{C}^{N_t}$ is a vector of transmitted symbols and $\mathbf{y} \in \mathbb{C}^{N_r}$ is a vector of received symbols. $N_t$ and $N_r$ represent the number of transmitted and received symbols, respectively. $\mathbf{n} \in \mathbb{C}^{N_r}$ is an i.i.d. complex Gaussian noise vector with covariance matrix $\sigma_n \mathbf{I}$. We assume that transmitted symbols are uniformly drawn from the discrete constellation $\mathcal{C}$ containing a number of $2^K$ complex symbols. In practice, $\mathcal{C}$ is the set of symbols in a quadrature amplitude modulation (QAM) constellation. $\mathbf{H}$ is the digital channel matrix between the transmitter and receiver, and $K$ is the modulation order.

Given channel knowledge (or an estimate of $\mathbf{H}$), the received vector $\mathbf{y}$, and the assumption that transmitted bits are chosen uniformly at random, the exact log-likelihood ratio (LLR) for the $i$-th bit of the $k$-th transmitted symbol is defined as \cite{sklar1997primer}
\begin{equation}
\label{eq:llr_probs}
    \Lambda_{i, k} = \log \frac{P(\mathbf{y} | b_{k,i}=1)}{P(\mathbf{y} | b_{k,i}=0)}.
\end{equation}

Under the i.i.d. Gaussian noise assumption, the optimal maximum-likelihood (ML) estimator for the LLR is expanded as
\begin{equation}
\label{eq:llr_def}
    \Lambda_{i, k}^{(\textrm{ML})} = \log \frac{\mathlarger{\sum\limits_{\mathbf{x} \in \mathcal{C}, b_{k, i} = 1}} \exp{- \frac{\norm{\mathbf{y} - \mathbf{H} \mathbf{x} }_2^2}{\sigma_n^2} }}
    {\mathlarger{\sum\limits_{\mathbf{x} \in \mathcal{C}, b_{k, i} = 0}} \exp{- \frac{\norm{\mathbf{y} - \mathbf{H} \mathbf{x} }_2^2}{\sigma_n^2} }}.
\end{equation}

The sums in \eqref{eq:llr_def} involve a number of $2^{KN_t} / 2$ terms for both the denominator and the numerator, leading to a prohibitive computational complexity at inference time even for moderate values of $N_t$ and $K$. It is an important research problem to develop greedy approaches that approximate the optimal LLR solution and achieve a low-latency and accurate inference algorithm.

Given the QR decomposition of $\mathbf{H} = \mathbf{Q}\mathbf{R}$ and exploiting the fact that $\mathbf{Q}$ is always a matrix with orthonormal columns and satisfies $\norm{\mathbf{Q} \mathbf{x}}_2 = \norm{\mathbf{x}}_2$ for any complex vector $\mathbf{x}$, \eqref{eq:llr_def} can be rewritten as
\begin{equation}
\label{eq:llr_def_qrspace}
    \Lambda_{i, k}^{(\textrm{ML})}  = \log \frac{\mathlarger{\sum\limits_{\mathbf{x} \in \mathcal{C}, b_{k, i} = 1}} \exp{- \frac{\norm{\hat{\mathbf{y}} - \mathbf{R} \mathbf{x} }_2^2}{\sigma_n^2} }}
    {\mathlarger{\sum\limits_{\mathbf{x} \in \mathcal{C}, b_{k, i} = 0}} \exp{- \frac{\norm{\hat{\mathbf{y}} - \mathbf{R} \mathbf{x} }_2^2}{\sigma_n^2} }},
\end{equation}

\noindent where $\hat{\mathbf{y}} = \mathbf{Q}^\mathrm{H} \mathbf{y}$. The upper triangular structure of $\mathbf{R}$ is exploited for an efficient implementation of the zero-forcing successive interference cancellation (ZF-SIC) detection algorithm \cite{kobayashi2016ordered}, where detection starts from the $N_t$-th stream and proceeds upwards, as described in Algorithm \ref{alg:zf-sic}.
\begin{algorithm}[!t]
  \begin{algorithmic}[0]
    \REQUIRE $\hat{\mathbf{y}}, \mathbf{R}, \mathcal{C}$
    \STATE // \textit{Initialize}
    \STATE $\mathbf{\Lambda}^{(\textrm{ZF-SIC})} \leftarrow \mathbf{0}_{K \times N_t},\tilde{\mathbf{x}} \leftarrow \mathbf{0}_{N_t \times 1}, \tilde{\mathbf{y}} \leftarrow \mathbf{0}_{N_t \times 1}$
    \FOR{$k = N_t:1:-1$}
        \STATE // \textit{Subtract previous interference}
        \STATE $\tilde{y}_k = \hat{y}_k - \sum\limits_{j=N_t}^{k+1} r_{k, j} \tilde{x}_j$
        \STATE // \textit{Estimate LLR by marginalization}
        \STATE $\Lambda_{i, k}^{(\textrm{ZF-SIC})} = \log \frac{\sum_{x_j \in \mathcal{C}, b=1} \exp -|\tilde{y}_k - r_{k, k} x_j|^2}
        {\sum_{x_j \in \mathcal{C}, b=0} \exp -|\tilde{y}_k - r_{k, k} x_j|^2} $
        \STATE // \textit{Update hard estimate}
        \STATE $\tilde{x}_k = \argmin_{x_j \in \mathcal{C}} |\tilde{y}_k - r_{k, k} x_j|^2 $
    \ENDFOR
    \ENSURE $\mathbf{\Lambda}^{(\textrm{ZF-SIC})}$
  \end{algorithmic}
  \caption{ZF-SIC Detection via QR Decomposition.}
  \label{alg:zf-sic}
\end{algorithm}

For the remainder of the paper, we assume that the LLR matrix $\mathbf{\Lambda}$ is represented as a vector of length $N_t K$. The task of LLR quantization consists in designing a pair of functions $f_\textrm{Q}, g_\textrm{Q}$ such that $g_\textrm{Q}(f_\textrm{Q}(\mathbf{\Lambda})) \approx \mathbf{\Lambda}$, and $f_\textrm{Q}(\mathbf{\Lambda})$ is represented using a finite number of bits. This is always the case for finite-precision arithmetic on digital computers, thus it is understood that we target a low number of bits when compared to floating-point representations. We note that the approximation is with respect to a suitably chosen distance metric, which we discuss in the sequel. The task of LLR estimation consists in designing a (potentially stochastic) function $f_\textrm{E}$ such that $f_\textrm{E}(\mathbf{y}, \mathbf{H}, \sigma_n) \approx \mathbf{\Lambda}$, where, again, the approximation is with respect to a metric discussed later. While we assume exact channel knowledge for the moment, we investigate the impact of impairments in Section \ref{sec:results}.

\section{Compressed Representation of ZF-SIC Solution}
\label{sec:theory}
To motivate our proposed deep learning approach, we prove a theorem concerning the induced structure of the ZF-SIC solution, $\mathbf{\Lambda}^{(\textrm{ZF-SIC})}$. Intuitively, the main takeaway of this section is that the sub-optimal soft-output ZF-SIC estimation algorithm outputs a solution that can be represented using a feature space with a number of dimensions lower than the total number of elements in $\mathbf{\Lambda}^{(\textrm{ZF-SIC})}$. Importantly, this number of dimensions is invariant to the modulation order $K$. We then conjecture that this compression ratio is optimal and posit that a data-driven approach can cover the performance gap between the ZF-SIC and ML estimators and learn a low-complexity, high-performance estimation algorithm that can also compress the returned solution.

To begin, we consider the soft-output version of the ZF-SIC algorithm given by Algorithm \ref{alg:zf-sic}. Given channel knowledge $\mathbf{H}$, the algorithm sequentially estimates transmitted symbols $x_i$ by leveraging the QR decomposition of $\mathbf{H}$. Once a symbol is estimated, it is subtracted from all other received streams and detection continues. Note that for the purposes of our theorem we do not assume any specific channel ordering, such as the one used in V-BLAST \cite{wolniansky1998v} and its extensions.

\begin{theorem}[Dimension of ZF-SIC Solution]
\label{theorem:zf-sic}
Let $\mathbf{\Lambda}^{(\textrm{ZF-SIC})} \in \mathbb{R}^{KN_t}$ be the solution obtained by the ZF-SIC algorithm. Then, there exists a surjective function $g:\mathbb{R}^{d_\textrm{ZF-SIC}} \mapsto \mathbb{R}^{KN_t}$, such that:
$$g(\mathbf{z}) = \mathbf{\Lambda}^{(\textrm{ZF-SIC})},$$
\noindent where
$$d_\textrm{ZF-SIC} = 3 N_t,$$
\noindent for all modulation orders $K$.
\end{theorem}

\begin{IEEEproof}
See Appendix A.
\end{IEEEproof}

Theorem \ref{theorem:zf-sic} states that any vector solution obtained by the ZF-SIC algorithm admits an \textit{exact} low-dimensional representation. Intuitively, this result can be made apparent by noting that the ZF-SIC algorithm effectively approximates the $\mathbf{R}$ matrix with its diagonalized version. In this case, all symbols are decoupled and the set of LLR values corresponding to a specific transmitted stream can be exactly stored using \textit{three} values. This leads to a total of $3N_t$ real values required to store the entire $\mathbf{\Lambda}$ vector. For a single-input single-output scenario, this is the basis for the prior work in \cite{arvinte2019deep}, where the vector of LLR values is compressed to a dimension of three, regardless of modulation size. For the purpose of our work, Theorem \ref{theorem:zf-sic} extends this to the solution output by the ZF-SIC estimation algorithm in arbitrary MIMO scenarios. While we assume infinite numerical precision is used in the function $g$, we verify in Section \ref{sec:results} that finite precision does not change the implications of Theorem \ref{theorem:zf-sic}.

While the above theorem does offer insight into the ZF-SIC algorithm, it does not clarify whether a similar conclusion holds for the ML solution as well. We make the following conjecture:

\begin{conj}[Optimality of $d_\textrm{ZF-SIC}$]
\label{conjecture}
There exists a surjective function $g:\mathbb{R}^{d_\textrm{ZF-SIC}} \mapsto \mathbb{R}^{KN_t}$ such that:
$$g(\mathbf{z}) \approx \mathbf{\Lambda}^{(\textrm{ML})},$$
\noindent for all modulation orders $K$.
\end{conj}

Note the usage of approximation instead of strict equality. We consider two LLR vectors approximately equal if the Hamming distance between the two decoded codewords asymptotically goes to zero as the signal-to-noise-ratio (SNR) increases, where SNR is defined as $\frac{\mathbb{E}[\norm{\mathbf{H}}_F^2]}{\sigma_n^2}$. In coded systems, this is equivalent to verifying that the block error rate decreases to zero at the same rate as the ML solution as the SNR increases when compression is applied.

Since there are a total $KN_t$ LLR values corresponding to a single channel use, the compression ratio of ZF-SIC is $R_\textrm{ZF-SIC} = \frac{KN_t}{d_\textrm{ZF-SIC}} = \frac{3}{K}$. If $K \geq 3$, then the solution of ZF-SIC \textit{always} admits a compressed representation. One can thus interpret the simple (in terms of computational complexity) ZF-SIC method as finding an implicitly compressed estimate of the ML solution, which in turn helps alleviate computational complexity through the low-dimensional assumption. Finally, note that our observations do not also cover the performance of algorithms that have a compression ratio lower than $R_\textrm{ZF-SIC}$. While it is entirely possible that such compression leads to an asymptotically decaying block error rates, our results in Section \ref{sec:results} show that this performance is no longer close to the ML algorithm, thus we do not consider them optimal.

\section{EQ-Net: Joint Estimation and Quantization}
\label{sec:eqnet}

The previous section motivates EQ-Net as a feature learning algorithm with a compressed latent feature representation. The size of this latent space is equal to the implicit dimension of the ZF-SIC estimate, but the desired performance is that of the ML estimate. Since such an algorithm does not have a closed-form expression, we resort to a data-driven approach for learning a model which achieves near-optimal end-to-end performance.

\begin{figure}[h]
\centering
\includegraphics[width=\linewidth,height=0.35\textheight,keepaspectratio]{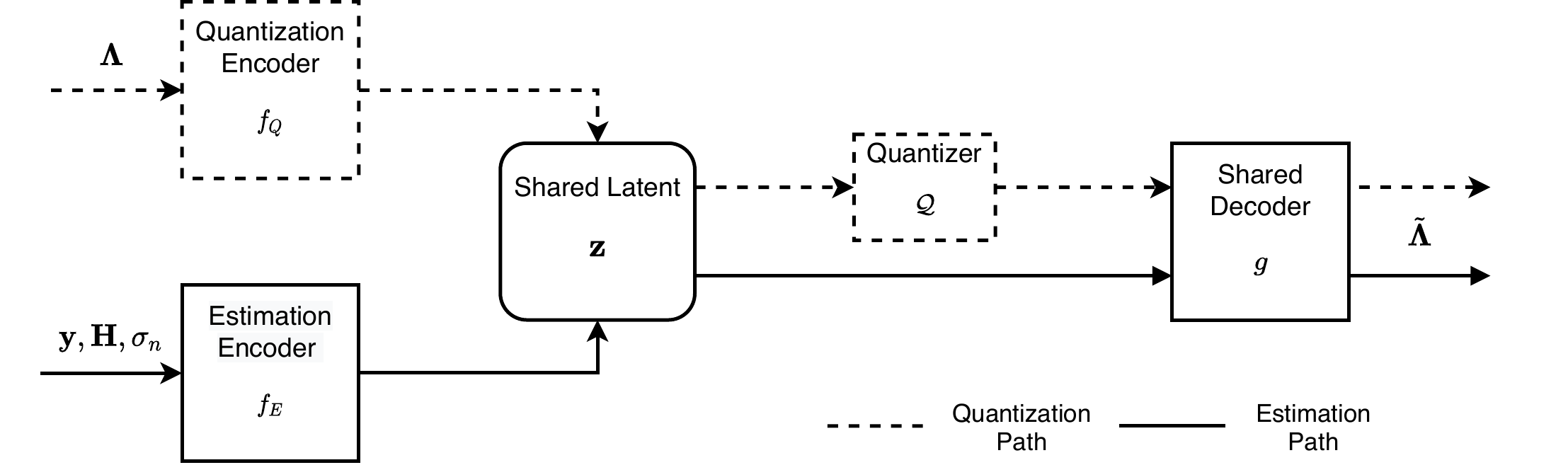}
\caption{High-level block diagram of the proposed architecture. The two encoders and the shared decoder are implemented as deep neural networks with a small number of hidden layers to ensure a low-latency signal path. The dotted lines indicate all components trained in the first stage and can be used for quantization at test time. The estimation encoder $f_\textrm{E}$ is trained \textit{after} the first stage, with all other components frozen.}
\label{fig:block_diagram}
\end{figure}

EQ-Net is a supervised method that uses compression as a \textit{pretext task} for estimation: when learning the estimator, we do not learn a direct mapping from the received symbols to the vector of LLR values, but rather split the learning in two separate stages. In the first stage, we train a compression encoder and decoder functions. Once the compression model has converged, we train another encoder that maps the pair formed by the received symbols and channel knowledge to the corresponding latent code from the first stage. The architecture of EQ-Net is comprised of three functional blocks and a high-level functional diagram is shown in Fig. \ref{fig:block_diagram}. Our ablation experiments in Section \ref{sec:results} show that this two-stage procedure is an essential step when training a model with limited depth and width and that single stage end-to-end training falls into unwanted local minima.

\subsection{Stage 1: Compression}
In the first stage, we train an autoencoder composed from the quantization encoder $f_\textrm{Q}$ and shared decoder $g$. The input to the autoencoder is a vector of LLR values estimated with the ML algorithm or, when this is not feasible even at training time, with an approximate ML algorithm such as the soft-output sphere detection algorithm. The loss function used is the same as \cite{arvinte2019deep}, namely the inter- and intra-LLR weighted mean squared error loss given by
\begin{equation}
\label{eq:wmse_training}
    L_{\textrm{Q}} = \sum_{i=1}^{K}w_i\frac{\norm{g_\textrm{Q} \circ \mathcal{Q} \circ f_\textrm{Q}(\Lambda_i) - \Lambda_i}_2^2}{|\Lambda_i| + \epsilon}.
\end{equation}

We use a value of $\epsilon = 10^{-6}$ to stabilize the loss function around low-magnitude values. The weights $w_i$ are proportional to the average magnitude of each LLR in the training dataset and are pre-determined. The quantizer $\mathcal{Q}$ is a function that maps the interval $[-1, 1]$ to a discrete and finite set of points $\mathcal{C}$ with a resolution of $N_b$ bits. During training, we replace the hard quantization operator with a differentiable approximation during training to obtain useful gradients. We follow previous work \cite{arvinte2019deep} and use a simple Gaussian noise model during training as the operator $\mathcal{Q}(x) = x + u$, where $u$ is drawn from $\mathcal{N}(0, \sigma_u)$ and $\sigma_u = 0.001$. To prevent the network from learning a trivial solution by amplifying the magnitude of the latent components, we use a hyperbolic tangent activation on the output layer of $f_Q$.

Once the autoencoder is trained we obtain the latent representation $\mathbf{z}$ by passing all training samples through the encoder $f_Q$ but not through the quantizer (or its differentiable approximation). We include $\mathbf{z}$ as a separate functional block in Fig. \ref{fig:block_diagram} since it is useful to make the connection with the theorem in the previous section: $\mathbf{z}$ defines the geometry of the low-dimensional manifold our estimation algorithm will operate on. In this sense, our scheme leverages compression as a pretext task for learning a well-shaped, robust manifold of the data.

\subsection{Stage 2: Estimation}
In the second stage, we train an estimation encoder $f_\textrm{E}$ to map received samples and channel state information to the same latent code output by the quantization encoder, that is, to map directly on the manifold given by $\mathbf{z}$. The supervised estimation loss is given by
\begin{equation}
    L_\textrm{E} = \norm{f_\textrm{E}(\mathbf{y}, \mathbf{H}, \sigma_n; \theta_{e}) - f_\textrm{Q}({\mathbf{\Lambda}})}_1.
\end{equation}

Intuitively, the interaction between the pair $f_\textrm{Q}$ and $f_\textrm{E}$ acts as a feature teacher-student \cite{hinton2015distilling} algorithm that performs knowledge distillation. Since we already have a good decoder in $g$, this allows us to give up the end-to-end reconstruction objective in the second stage and map received symbols to a static, pre-trained latent space. In our experiments, we show that this is critical for the convergence of shallow, low-latency networks to high performance solutions.

\subsection{Operating Modes}
At test time, EQ-Net inference is executed independently for each subcarrier in an OFDM system. This follows the lines of prior work in deep learning-based MIMO detection schemes \cite{he2019model,sholev2020neural} and allows for compact models. Given a set of received antenna symbols $\mathbf{y}$, an estimated channel $\mathbf{H}$ and estimated noise variance $\sigma_n^2$, our model uses them to operate in one of the following modes.

In \textit{quantization} mode our algorithm is compatible with any external module that estimates the log-likelihood ratios. The vector of values corresponding to a channel use is fed to the encoder $f_Q$ and the result is quantized with the operator $\mathcal{Q}$, yielding a bit array representing the quantized LLR vector. The low-resolution bit array is stored for future use or efficiently transmitted over a secondary channel, after which the recovered LLR vector is obtained by applying $g$. In \textit{estimation} mode the algorithm directly passes $\mathbf{y}, \mathbf{H}$ and $\sigma_n$ through the estimation encoder and the decoder without quantizing the latent representation. This produces the estimated LLR vector $\tilde{\mathbf{\Lambda}}$. 

\subsection{Implementation Details}
\begin{figure*}[!t]
\centering
\subfloat[Quantization encoder $f_\mathrm{Q}$ with a simple and shallow MLP structure.]{\includegraphics[width=0.3\linewidth]{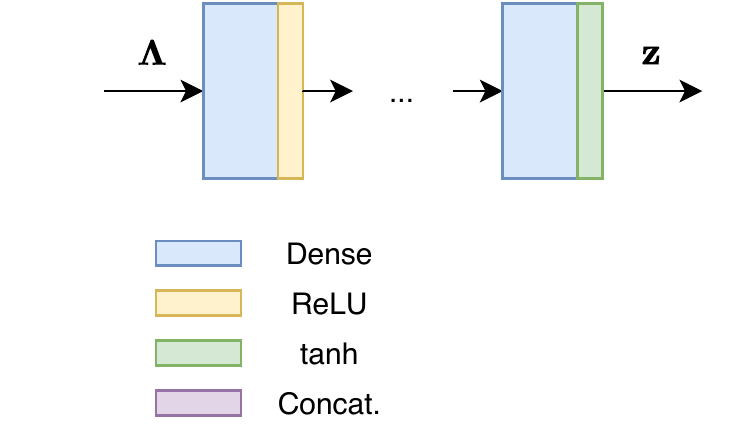}\label{subfig:fQ}}
\hfil
\subfloat[Shared decoder $g$ with a branched architecture.]{\includegraphics[width=0.3\linewidth]{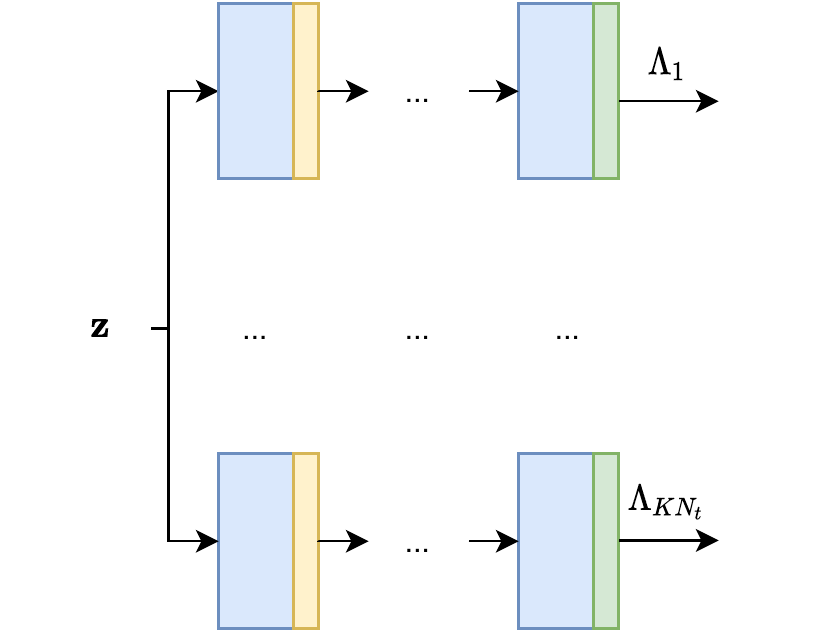}\label{subfig:g}}
\hfil
\subfloat[Input layers and one residual block of the estimation encoder $f_\mathrm{E}$. Multiple residual blocks can be concatenated to form $f_\mathrm{E}$.]{\includegraphics[width=0.35\linewidth]{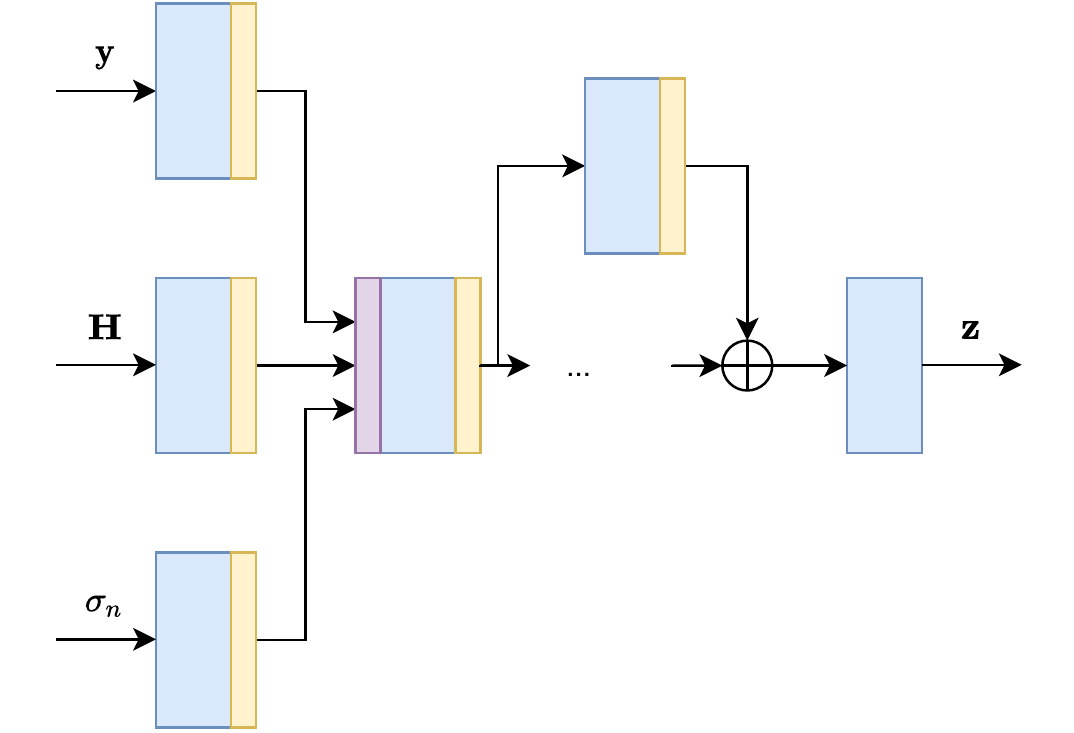}\label{subfig:fE}}
\caption{Internal architectures for each of the functional blocks in Fig. \ref{fig:block_diagram} implemented as deep neural networks.}
\label{fig:detailed_blocks}
\end{figure*}

The models $f_\mathrm{Q}$, $f_\mathrm{E}$ and $g$ are all comprised of deep neural networks, sharing the common design principle of reducing the end-to-end latency as much as possible to make them suitable for fast inference. The most straightforward way of achieving this is by limiting the depth of all networks while potentially increasing their width. Conversely, depth can be increased at the cost of latency to improve system performance. A detailed block diagram of the models is shown in Fig. \ref{fig:detailed_blocks}. All models use the ReLU activation function across all hidden layers. Across training and testing, the LLR vector is pre-processed by converting it to the hyperbolic domain via the element-wise function $f(x) = \tanh{(x/2)}$. 

The quantization encoder $f_Q$ has a simple structure as a fully-connected multi-layer perceptron (MLP) with six hidden layers, each with a width of $4N_tK$, and an output layer of size $d_\textrm{ZF-SIC} = 3N_t$. Note that the width scales with modulation order but the depth is fixed, allowing us to increase the expressive power of the network without sacrificing latency in higher-order modulation scenarios.

The estimation encoder $f_E$ takes as input the triplet $(\mathbf{y}, \mathbf{H}, \sigma_n)$ and passes their flattened, real-valued (obtained by concatenating the real and imaginary parts) versions through a separate dense layer for each channel, followed by a concatenation operation. This is analogous to an early feature fusion strategy, since the three input signals have different physical dimensions and interpretations. A single block of the estimation encoder contains an additional six hidden layers with residual connections between them, as shown in Fig. \ref{subfig:fE}. The reason for not including these connections in the quantization encoder is empirical and based on the satisfactory performance offered by $f_Q$ without them.

The shared decoder $g$ uses the same branched architecture described in \cite{arvinte2019deep}: the latent representation is separately processed on $N_t \times K$ parallel multi-layer perceptron (MLP) networks, each with six hidden layers. This architecture learns a separate decoding function for each individual entry in the log-likelihood ratio vector, while still taking the entire latent representation as input.

To discretize the latent representation during testing, where gradients are no longed needed, we learn a factorized quantization codebook by separately applying a quantization function to each dimension of the latent space. We use the same data-based approach as \cite{arvinte2019deep} and train a k-means++ \cite{arthur2006k} scalar quantizer after $f_Q$ and $g$ have been trained. Note that this is not the only possible choice here, but in practice we find little benefit to training a vector quantizer.

\section{Experimental Results}
\label{sec:results}
For training, we use the Adam optimizer \cite{kingma2014adam} with a batch size of $32768$ samples, a learning rate of $0.001$ and default Tensorflow \cite{abadi2016tensorflow} parameters for both stages. Training data consists of pre-generated ML estimates of LLR values from $10000$ packets at six logarithmically spaced SNR values, such that the block error rate of ML estimation is between $1$ and $0.0001$. We use a low-density parity check (LDPC) code of size $(324, 648)$, leading to a total of $2.7$ million training samples for a $2$-by-$2$ $64$-QAM scenario and slightly less for $4$-by-$4$ $16$-QAM. We reserve $20\%$ of these samples as validation data. Testing is performed with the same channel coding scheme across a wider array of SNR values. The scheme does not suffer a performance loss if tested on other codes (e.g., polar), but we omit these results for brevity.

\subsection{Verifying Theorem 1}
To verify Theorem 1 and Proposition 1, we train a series of EQ-Net models, where the only parameter that changes is the bottleneck size. We do not apply any numerical quantization, and assume that the ML estimate is already known, focusing this experiment only on verifying our theoretical claims. For exemplification, we target a $2$-by-$2$, $64$-QAM scenario under i.i.d. Rayleigh fading, but this result holds for arbitrary configurations. For this scenario, we have that $d_\textrm{ZF-SIC} = 6$, thus Theorem 1 informs us that the simplest possible (and with the worst non-trivial performance) algorithm, ZF-SIC, implicitly projects the log-likelihood ratio vector onto a \textit{six}-dimensional manifold.

\begin{figure}
\centering
\includegraphics[width=\linewidth,height=0.35\textheight,keepaspectratio]{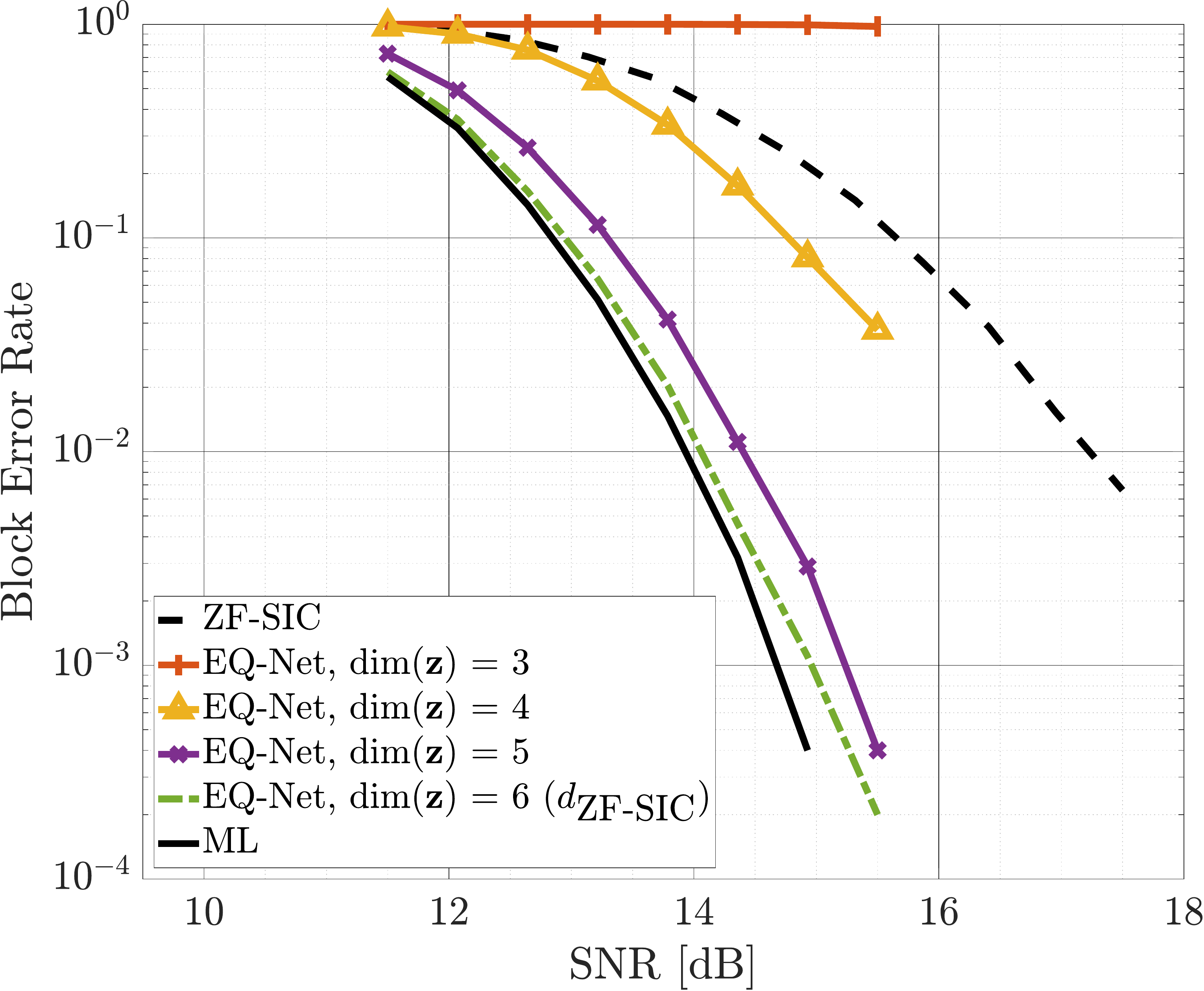}
\caption{Impact of bottleneck dimension on different instances of EQ-Net and the two reference algorithms without any numerical quantization in an i.i.d. Rayleigh fading, $2 \times 2$ $64$-QAM scenario with an $(324, 648)$ LDPC code. A bottleneck size of six corresponds to the implicit dimension of the ZF-SIC solution (Theorem 1), but retains near-ML performance.}
\label{fig:latent_dim}
\end{figure}

Fig. \ref{fig:latent_dim} plots performance when the bottleneck size is the only parameter that is varied in a series of EQ-Net models. In all cases, the LLR vector is estimated with the optimal ML algorithm and is then passed through the bottleneck, but not numerically quantized. For reference, we also plot the two conventional ZF-SIC and ML algorithms. Two conclusions can be drawn from this plot:
\begin{enumerate}
    \item A purely data-based approach \textit{can} find nonlinear, low-dimensional projections that improve the performance compared to the ZF-SIC algorithm and retain the same compression ratio.
    \item Any attempt to further compress the optimal LLR vector is met with an increase of error and departure from ML performance, but can still outperform ZF-SIC estimation.
\end{enumerate}

For the rest of the experiments, we use a latent space with $\textrm{dim}(\mathbf{z}) = 3N_t$ --- the same as the ZF-SIC algorithm --- due to the minimal performance loss incurred and to compress as much as possible. This has the benefit of a lower compression ratio as modulation size increases. While the second observation could be used to further compress the vector of log-likelihood ratios in higher-dimensional MIMO transmissions, where the performance/complexity gap between ML and ZF-SIC increases, the rest of our experiments are carried out on digital communication scenarios with $N_t \leq 4$ and an increased modulation size.

This result allows for the following interpretations of EQ-Net: \textit{i)} EQ-Net is a compression algorithm that retains near-ML performance while mapping the vector of LLR values to a bottlenecked feature space of size $d_\textrm{ZF-SIC}$ and \textit{ii)} EQ-Net is an estimation algorithm that encompasses ZF-SIC as a local minima in the loss landscape, but benefits from a sufficient number of trainable parameters to learn better performing solutions.

\subsection{Importance of Two-Stage Training}
\begin{figure}
\centering
\includegraphics[width=\linewidth,height=0.35\textheight,keepaspectratio]{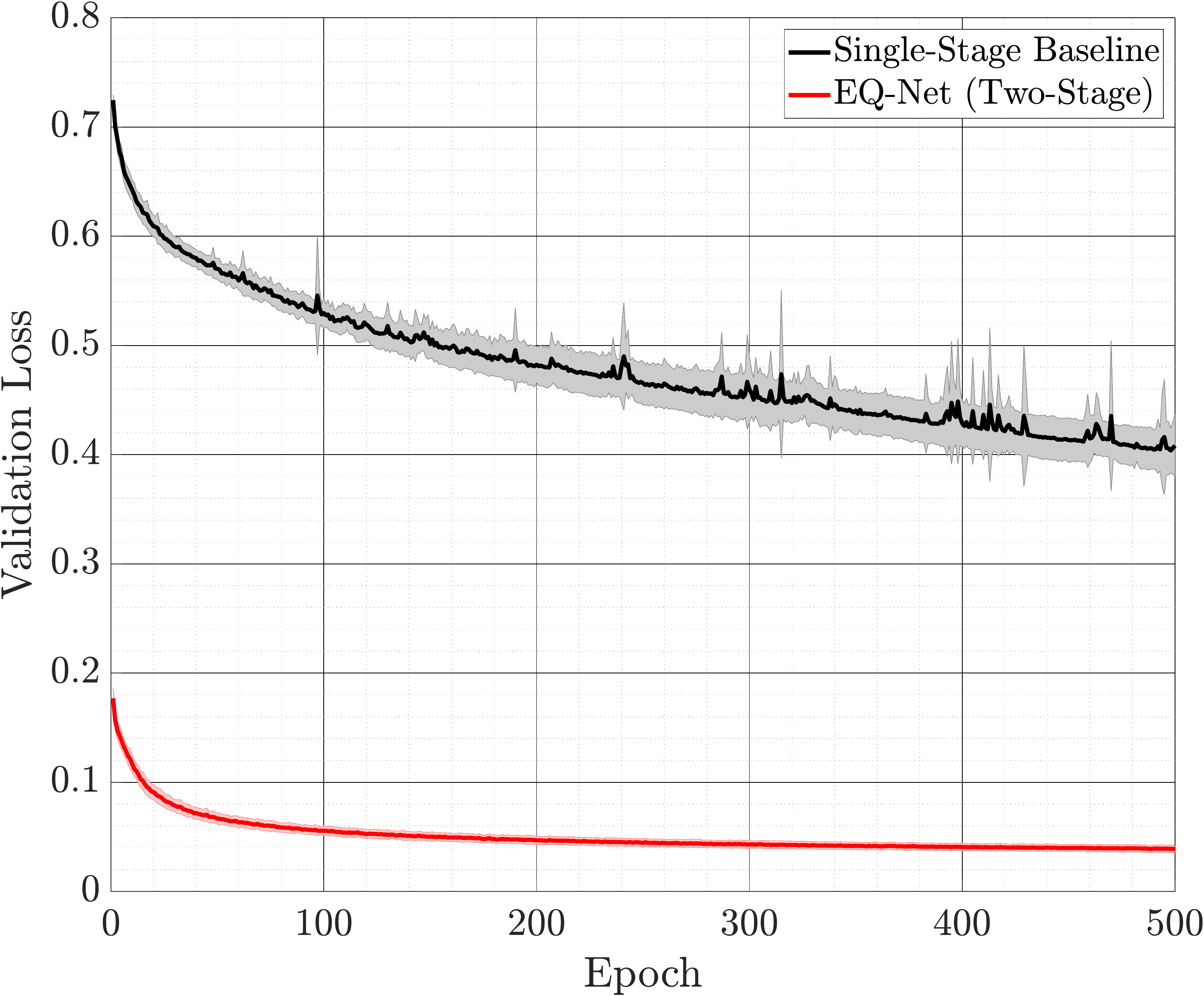}
\caption{Validation loss during training progression for EQ-Net and a na\"ive supervised baseline approach that jointly trains the estimation encoder and the decoder. In both cases, the networks have the same architecture and initial weights. All weights receive exactly the same number of gradient updates in the two experiments. The plot includes shaded standard deviation areas over ten runs.}
\label{fig:two_stage_training}
\end{figure}

To highlight the importance of quantization pre-training, we investigate the difference in performance against a baseline single-stage training method. To perform a calibrated comparison, we use exactly the same architecture in both cases. For the single-stage baseline, we train the two networks jointly for a total of $500$ epochs, with the learning rate divided by a factor of two if the validation loss does not improve for $100$ consecutive epochs. For EQ-Net, each stage is trained for the same number of $500$ epochs. We increase the starting learning rate for the joint approach to compensate for the untrained latent representation. Fig. \ref{fig:two_stage_training} plots the evolution of the validation loss across the training period for both methods in a $2$-by-$2$, $64$-QAM scenario under i.i.d. Rayleigh fading. This highlights the superior performance of the proposed method: baseline single-stage training is unstable and consistently falls into a local minima, whereas the two-stage method converges to a good solution, even though each component is separately trained for \textit{half} the time of joint training. Note that for SISO scenarios (more broadly, MIMO channels with orthogonal channel matrices), a two-layer neural network can be easily trained to match the performance of our approach \cite{shental2019machine}. This is owed to the particular structure of QAM with Gray coding, but this observation does not straightforwardly extend to high-dimensional channels.

\subsection{Estimation Performance}
We evaluate two variants of EQ-Net: EQ-Net-L (a low-latency architecture with one residual block) and EQ-Net-P (a high-performance architecture with three residual blocks). We compare our method with two state-of-the-art deep learning baselines: the scheme in \cite{sholev2020neural}, which we further refer to as NN-Det for brevity, and the OAMP-Net2 approach in \cite{he2019model}.

We implement both baselines in Tensorflow and follow the original design principles as closely as possible, while searching for their best hyperparameters. For $2$-by-$2$ $64$-QAM, we train two variants of NN-Det: NN-Det-P ($4$ unfolded blocks) with the same end-to-end performance as our proposed method and NN-Det-L ($3$ unfolded blocks) which trades off some of the performance for lower end-to-end latency. For $4$-by-$4$ $16$-QAM, we train a single NN-Det model with $10$ unfolded blocks, labeled as high-performance. For OAMP-Net2, we only use the detector part with trainable step sizes and do not perform channel estimation, instead relying on exact CSI. At test time, for both NN-Det and OAMP-Net2 we form the log-likelihood ratios by summing all the corresponding symbol probabilities for the terms in \eqref{eq:llr_def}, since these methods output symbol probabilities. Exact implementations of the baselines are available along with our source code.


\begin{figure}
\centering
\includegraphics[width=\linewidth,height=0.35\textheight,keepaspectratio]{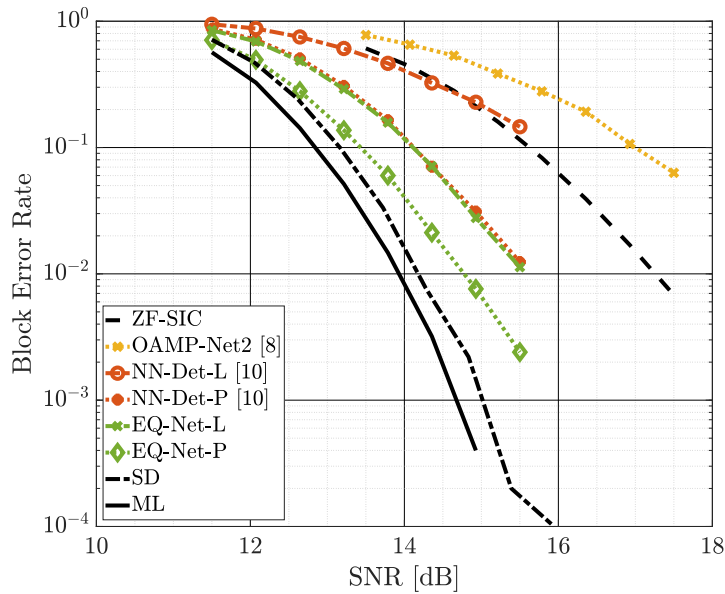}
\caption{Estimation (MIMO detection) performance of EQ-Net versus state-of-the-art approaches in a $2$-by-$2$, $64$-QAM Rayleigh fading and LDPC-coded scenario. Table \ref{table:latency_qam64} further compares the end-to-end latency of all methods.}
\label{fig:estimators_qam64}
\end{figure}

\begin{table}[!t]
\renewcommand{\arraystretch}{1.3}
\caption{Latency benchmarks of evaluated soft-output MIMO detection algorithms for a $2$-by-$2$ $64$-QAM scenario. All values expressed in milliseconds. $B$ represents the batch size during inference.}
\label{table:latency_qam64}
\centering
\begin{tabular}{|p{1.9cm}|c|c|c|c|c|c|}
\hline
& EQ-Net-L & EQ-Net-P & NN-Det-L & NN-Det-P & OAMP-Net2 & Sphere Dec. \\
\hline
Latency CPU, $B = 16$ & $1.767$ & $2.292$ & $2.565$ & $3.073$ & $2.376$ & $4.547$ \\
\hline
Latency CPU, $B = 1$ & $1.511$ & $1.903$ & $2.269$ & $2.689$ & $2.125$ & $0.6935$ \\
\hline
\hline
Latency GPU, $B = 8192$ & $6.523$ & $8.767$ & $12.61$ & $15.91$ & $25.94$ & --- \\
\hline
\end{tabular}
\end{table}

We measure end-to-end latency by implementing all methods as compact computational graphs in Keras + Tensorflow 1.15 and  timing the average duration of $10000$ calls of the \texttt{predict} method (after a number of warm-up rounds). We measurement execution times of the model calls using the \texttt{timeit} Python module. The CPU is an Intel i9-9900x with 10 cores running at $3.5$ GHz and the GPU is an NVIDIA RTX 2080Ti. 

We also include the performance and latency profile of the Sphere Decoding (SD) algorithm as a baseline high-performance, high-latency algorithm. We use the default MATLAB implementation based on the algorithm in \cite{studer2008soft}, which uses a single tree traversal and natively supports batch decoding, and report the average execution time across ten SNR values. From Tables \ref{table:latency_qam64} and \ref{table:latency_qam16}, it is clear that, while SD is efficient for a batch size of $1$, the degree of parallelism is much lower, explained by the heavy use of sorting and the cost of QR decomposition. This observation also holds for ZF-SIC. In contrast, all deep learning-based approaches do not involve any sorting operations, leading to more efficient parallel implementations.

\begin{figure}
\centering
\includegraphics[width=\linewidth,height=0.35\textheight,keepaspectratio]{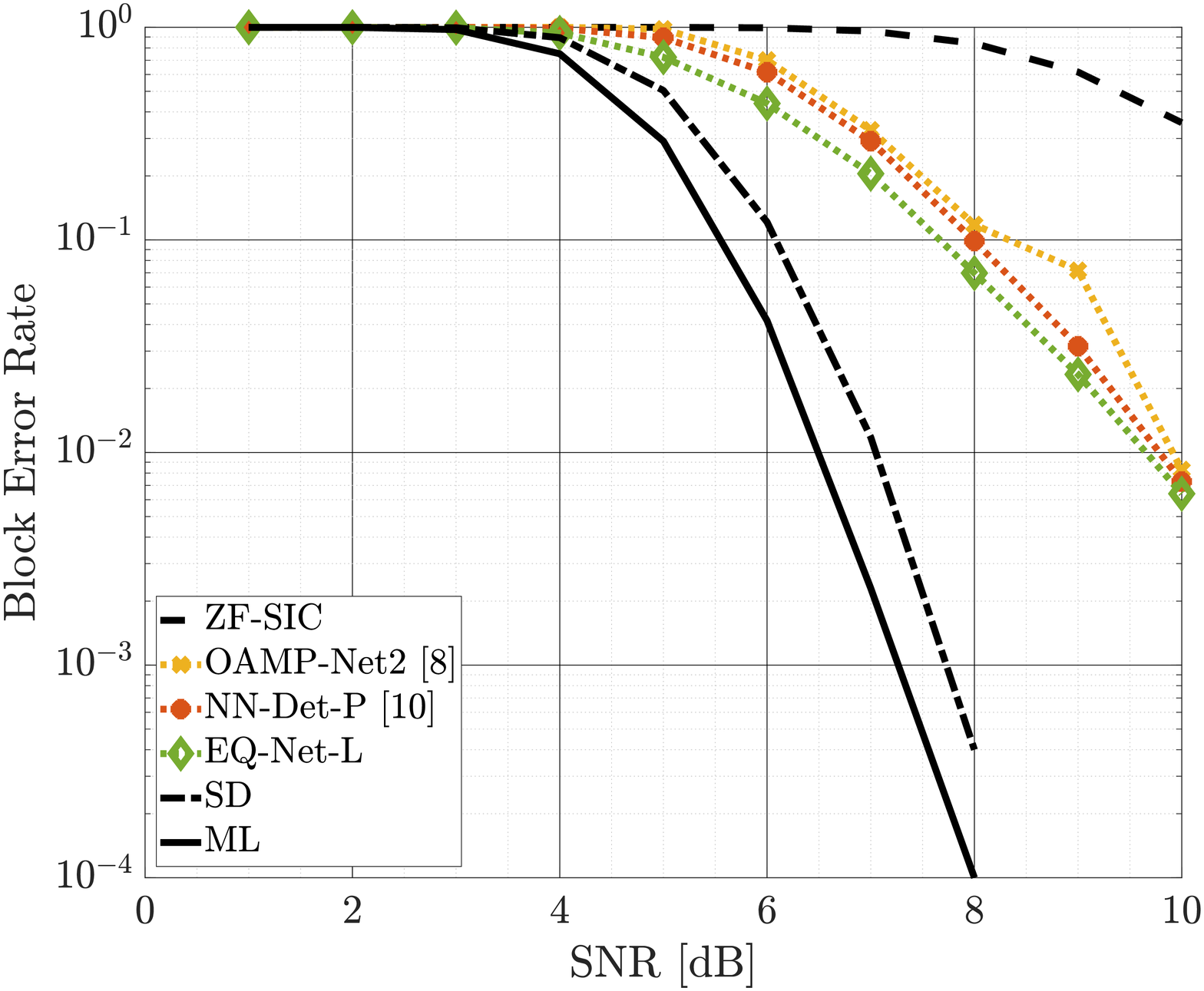}
\caption{End-to-end estimation performance of proposed algorithms against state-of-the-art approaches in a $4$-by-$4$, $16$-QAM Rayleigh fading scenario. Table \ref{table:latency_qam16} further compares the end-to-end latency of all methods.}
\label{fig:estimators_qam16}
\end{figure}

We find that both variants of EQ-Net surpass the prior work in performance with a gain of more than $2$ dB and in latency with at least a $60\%$ speed up. Fig. \ref{fig:estimators_qam64} and \ref{fig:estimators_qam16} plot the performance of the considered algorithms, while Tables \ref{table:latency_qam64} and \ref{table:latency_qam16} show the corresponding latency. For the $64$-QAM scenario, the low latency version of EQ-Net achieves the same performance as the high-performance (and latency) version of NN-Det. This is better highlighted in Fig. \ref{fig:estimators_qam16}, where we only compare EQ-Net-L with NN-Det-P and reach the same conclusion. This result highlights the benefits of pre-training the compressed feature space: while the number of learnable parameters in EQ-Net and NN-Det are approximately the same, the estimator learned by EQ-Net is much more efficient in using these parameters.

\begin{table}[!t]
\renewcommand{\arraystretch}{1.3}
\caption{Latency benchmarks of evaluated soft-output MIMO detection algorithms for a $4$-by-$4$ $16$-QAM scenario. All values expressed in milliseconds. $B$ represents the batch size during inference.}
\label{table:latency_qam16}
\centering
\begin{tabular}{|p{1.9cm}|c|c|c|c|}
\hline
& EQ-Net-L & NN-Det-P & OAMP-Net2 & Sphere Dec. \\
\hline
Latency CPU, $B = 16$ & $1.711$ & $5.957$ & $3.437$ & $19.52$ \\
\hline
Latency CPU, $B = 1$ & $1.561$ & $5.113$ & $3.154$ & $1.622$ \\
\hline
\hline
Latency GPU, $B = 8192$ & $6.523$ & $29.310$ & $35.727$ & --- \\
\hline
\end{tabular}
\end{table}

The reduced latency compared to the other deep learning methods can be attributed to the fact that the baseline approaches require additional linear algebra computations which are sequential in nature, such as matrix inversion and conjugate-multiply after each iteration. In contrast, our approach does not involve any of these steps: we take in the raw inputs and directly output the LLR vector. This fact becomes even more apparent when the algorithms are run on specialized inference hardware (i.e., GPUs). The last line in both tables highlights the increasing latency gap when the algorithm is deployed in a scenario with a massive number of users, where a single base station (or cloud computing device) performs computations in large batches.

\subsection{Quantization Performance}
For the quantization task, we investigate the performance of EQ-Net against the maximum mutual information quantizer in \cite{winkelbauer2015quantization} and the deep learning based scheme in \cite{arvinte2019deep}, which is designed for SISO scenarios. To extend this to a MIMO scenario, we simply split the LLR vector into sub-vectors along the transmitted stream and quantize each of them separately. This serves as a strong baseline that also reveals the gain obtained by considering redundancy across the spatial dimension. For our approach, we train a separate k-means++ quantizer with $64$ levels for each scalar dimension of the latent space, thus requiring six bits of storage for each latent component. For a $2$-by-$2$ $64$-QAM scenario, this amounts to compressing the entire vector of $12$ LLR values down to a $36$-bit codeword, leading to an effective compression ratio of three bits per LLR.

\begin{figure}
\centering
\includegraphics[width=\linewidth,height=0.35\textheight,keepaspectratio]{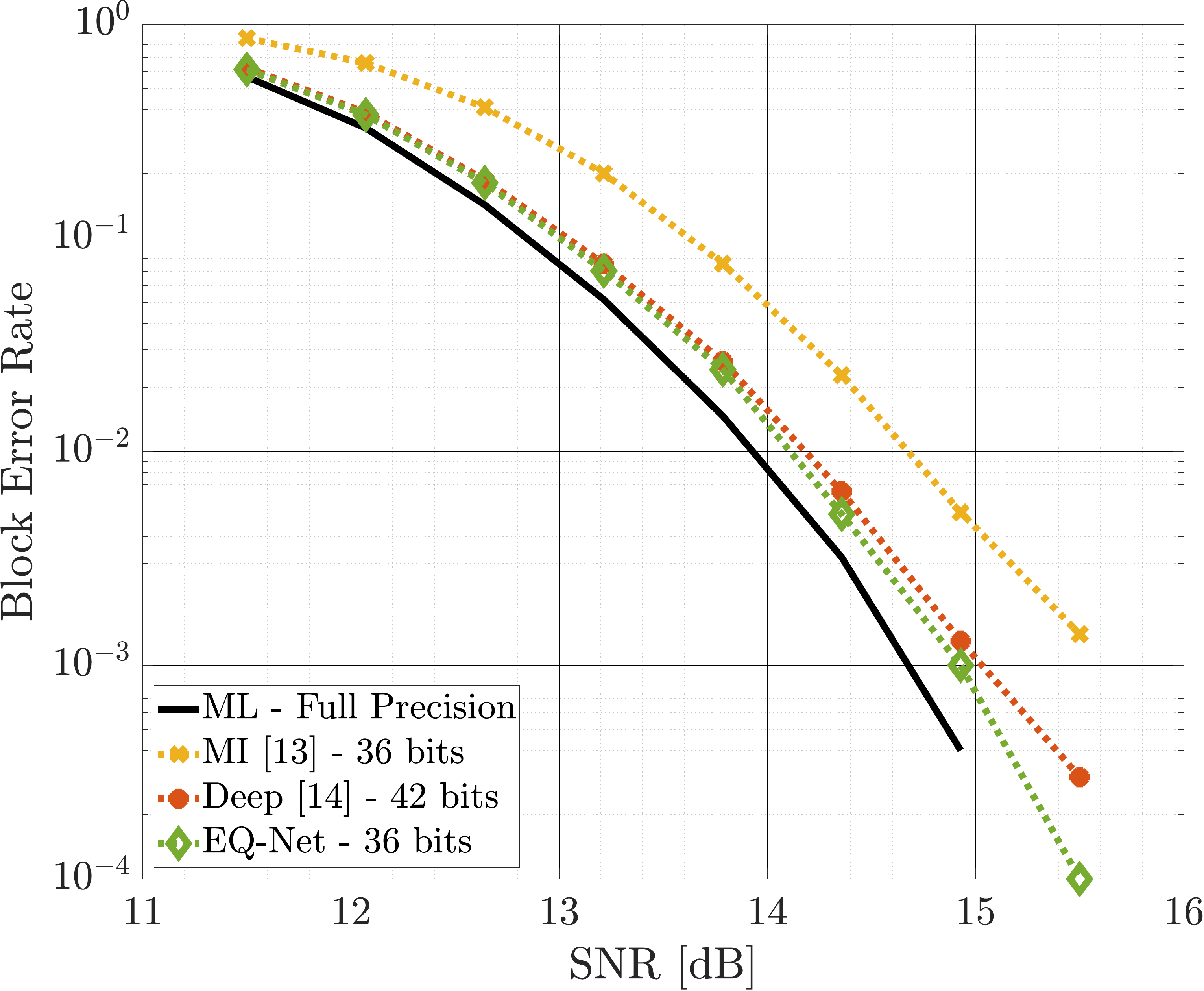}
\caption{Quantization performance of EQ-Net evaluated against state-of-the-art methods in a $2$-by-$2$, $64$-QAM, i.i.d. Rayleigh fading and LDPC-coded scenario.}
\label{fig:quantizers_qam64}
\end{figure}

The results in Fig. \ref{fig:quantizers_qam64} show that EQ-Net is superior to both baseline methods and can efficiently compress the LLR vector using a storage cost of $3$ bits per LLR with a minimal performance loss. Compared to \cite{arvinte2019deep}, EQ-Net achieves a $16\%$ compression gain with the same end-to-end performance, whereas compared to \cite{winkelbauer2015quantization} EQ-Net boosts the performance of the system $0.65$ dB while using same compression ratio. The difference in performance when compared to \cite{arvinte2019deep} highlights the importance of jointly learning a feature space across the spatial dimensions of MIMO channels.

EQ-Net also has the advantage of supporting a non-uniform quantization rate. For example, if five bits are used to store each latent component then the entire LLR vector can be stored using $30$ values with an effective $2.5$ bits per LLR. In contrast, classical compression methods that are applied to each LLR value individually would have to use an imbalanced compression scheme to achieve fractional rates, potentially leading to increased error floors in the system performance.

\subsection{Robustness to Distributional Shifts}
All the data-driven approaches so far have been tested on the same distributions as the ones they were trained on. Practical scenarios may, however, involve either severe distributional shifts (e.g., completely different fading models) or may be faced with imperfect channel state information during deployment. For LLR estimation and quantization, the most fragile part of the system is given by the CSI matrix $\mathbf{H}$ and any potential distributional mismatch that may occur. To evaluate the robustness of our approach, we consider models trained on $\mathbf{H}$ matrices from an i.i.d. Rayleigh fading model and tested --- without \textit{any} further adjustments or fine-tuning --- on realizations of the CDL-A channel model adopted by the 5G-NR specifications \cite{cdla}. This is a realistic model that has a larger degree of spatial correlation than the Rayleigh model and is driven by physical propagation laws.

\begin{figure}
\centering
\includegraphics[width=\linewidth,height=0.35\textheight,keepaspectratio]{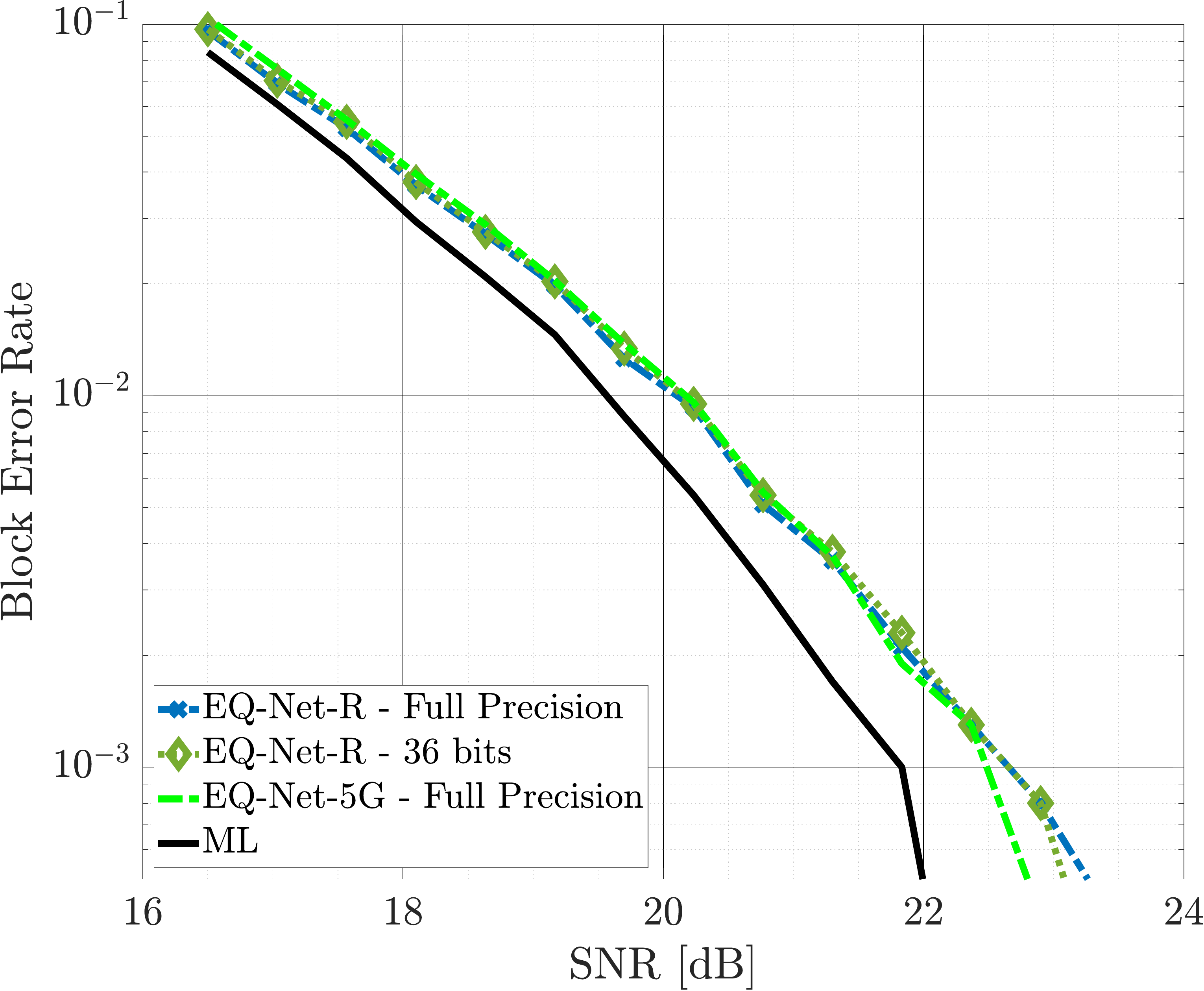}
\caption{Quantization performance of EQ-Net under a severe distributional shift induced by training on an i.i.d. Rayleigh fading channel and testing on the 5G-NR CDL-A MIMO channel model for a $2$-by-$2$ $64$-QAM LDPC-coded scenario.}
\label{fig:quantizers_robust_qam64}
\end{figure}

Fig. \ref{fig:quantizers_robust_qam64} shows the quantization performance under this shift in a $2$-by-$2$ $64$-QAM scenario, where EQ-Net-R stands for models trained on Rayleigh channels and tested on CDL-A. To accurately evaluate robustness, we also train a EQ-Net-5G model on realizations of the CDL-A channel model. While a performance gap of around $1$ dB is apparent between EQ-Net-5G and the ML solution, the EQ-Net-R scheme is overall extremely robust and does not exhibit any error floors, or deviations from performance. Furthermore, when quantizing using to the same ratio of $3$ bits per LLR, there is virtually no performance loss incurred. Thus, we conclude that the quantizers learned from the two different datasets are similar and applicable to a wide array of channel models.

\begin{figure}
\centering
\includegraphics[width=\linewidth,height=0.35\textheight,keepaspectratio]{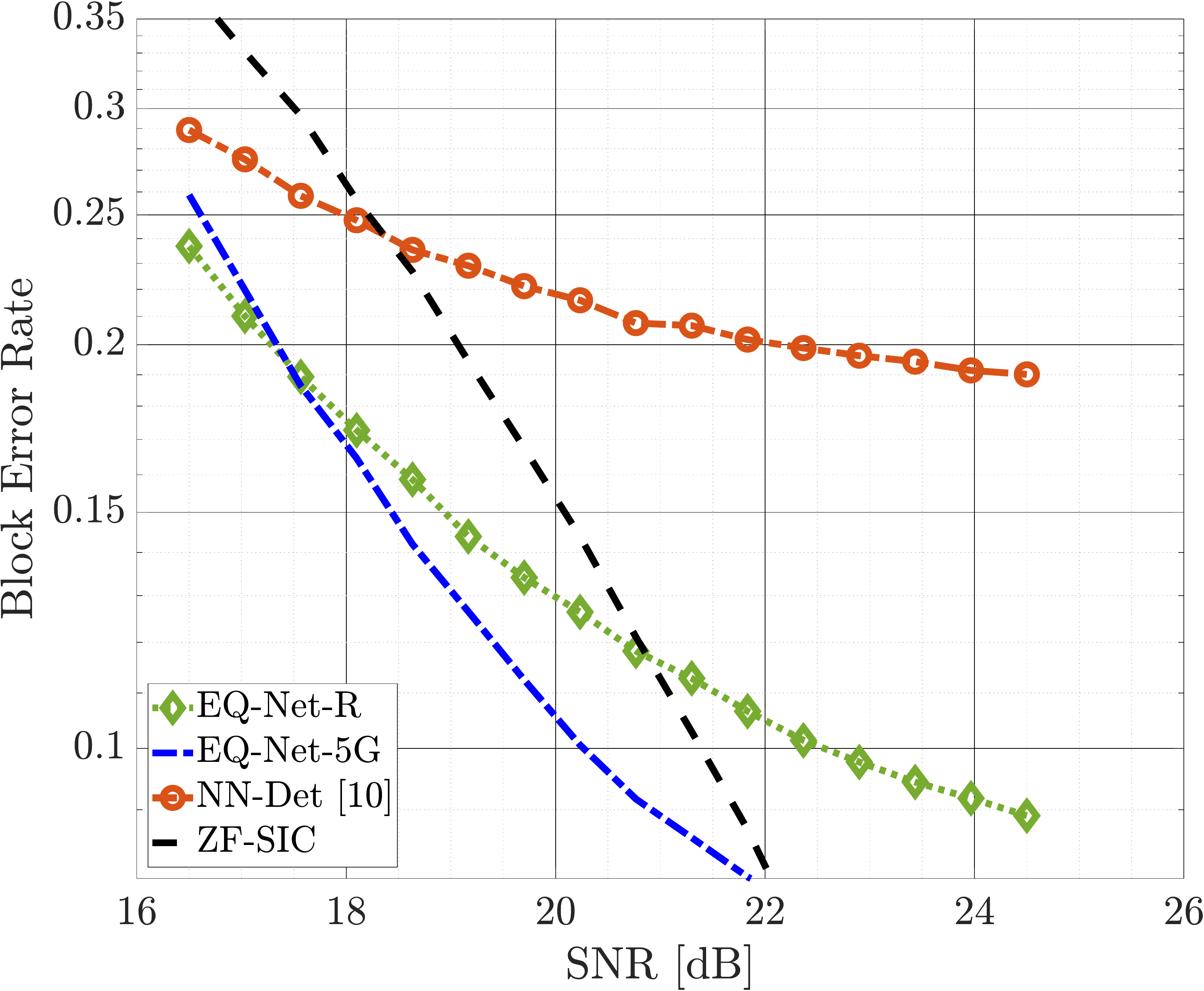}
\caption{Estimation performance of EQ-Net and the similarly performing NN-Det-P \cite{sholev2020neural} under severe distributional shift induced by training on an i.i.d. Rayleigh fading channel and testing on the 5G-NR CDL-A MIMO channel model for a $2$-by-$2$ $64$-QAM LDPC-coded scenario.}
\label{fig:estimators_robust_qam64}
\end{figure}

\begin{figure}
\centering
\includegraphics[width=\linewidth,height=0.35\textheight,keepaspectratio]{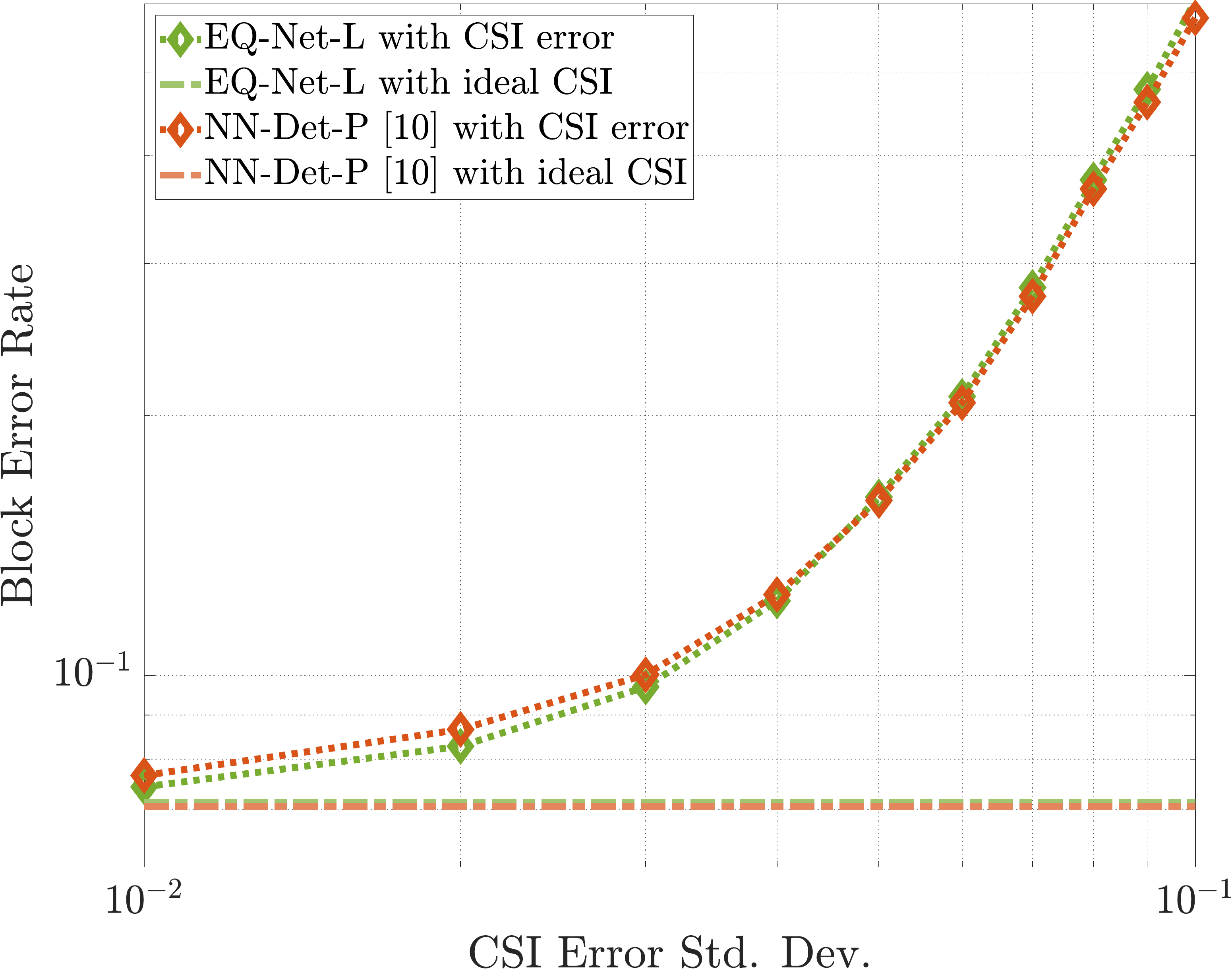}
\caption{Estimator robustness of EQ-Net and the similarly performing NN-Det-P \cite{sholev2020neural} under imperfect CSI for a $2$-by-$2$ $64$-QAM LDPC-coded scenario. For better visualization, we only plot the evolution of block error rate at an SNR value of $14.35$ dB as the CSI error increases. The two horizontal lines show the average end-to-end performance of the two algorithms with ideal CSI knowledge at the same SNR value.}
\label{fig:csi_qam64}
\end{figure}

Fig. \ref{fig:estimators_robust_qam64} investigates the estimation performance under the same shift and reveals a higher degree of robustness compared to the baseline NN-Det approach, retaining a performance close to that of the ZF-SIC algorithm. While the two methods started from a very similar same end-to-end performance, as shown in Fig. \ref{fig:estimators_qam64}, EQ-Net can avoid a severe error floor. We attribute this robustness to the bottleneck applied during training, as well as the particular expression of the WMSE loss used during training. Eq. \ref{eq:wmse_training} increases the weight of uncertain LLR values, which are more likely to occur in scenarios with ill-conditioned channels, such as those from the CDL-A model. Overall, our result highlights the potential of \textit{robust} feature learning for LLR estimation versus model-based approaches, which incorporate structural assumptions during training.

Finally, we also investigate the performance of our approach in the case of CSI estimation impairments at the receiver. For this we use a corrupted version of $\mathbf{H}$ coming from an i.i.d. model as $\hat{\mathbf{H}} = \mathbf{H} + \mathbf{N}$, where $\mathbf{N}$ is an i.i.d. Gaussian noise with covariance matrix $\sigma_{CSI} \mathbf{I}$. This models impairments coming from the channel estimation module. We choose a specific SNR value such that the two algorithms start from the same end-to-end performance. Fig. \ref{fig:csi_qam64} plots average performance over $10000$ channel realization with corrupted CSI and shows that our algorithm is robust to these impairments as well, remaining on par with the NN-Det algorithm and still benefiting from the much lower end-to-end latency in Table \ref{table:latency_qam64}.

\section{Discussion and Conclusions}
\label{sec:conc}
In this paper we have proposed a deep learning framework that jointly solves the tasks of log-likelihood ratio estimation and quantization in a MIMO scenario. We have used theoretical results on the zero-forcing with successive interference cancellation algorithm to design the dimension of a learned feature space. This insight has revealed that the solution of the ZF-SIC algorithm admits a lower-dimensional implementation and motivated our conjecture that this dimension leads to optimal compression. Our approach has been shown to be practical in terms of latency and is compatible with any MIMO system, such as the MIMO-OFDM used in 5G scenarios, relaying scenarios, or distributed communication systems, which would benefit from both quantization and estimation gains. Throughout evaluation, our approach has shown superior performance, distributional robustness and on-par impairment robustness to state-of-the-art estimation methods. 

One drawback that remains is the presence of an error floor when faced with severe distributional shifts at test time, as per Fig. \ref{fig:estimators_robust_qam64}. Even though our results show that this floor is much lower than that of prior work, there is still room for improvement, in at least overcoming the ZF-SIC algorithm across the entire SNR range. For example, our method could easily be extended to account for perturbations during training or be trained on a dataset that pools together realizations of multiple channel models. Another promising direction for future work is removing the requirement of training a separate model for each MIMO configuration and leveraging flexible deep learning architecture to learn truly universal algorithms for LLR processing.

\bibliographystyle{IEEEtran}
\bibliography{mybib_v2}

\begin{appendices}

\section{Proof of Theorem \ref{theorem:zf-sic}}
We prove this by induction. Consider the form of \eqref{eq:system} after left-side multiplication with $\mathbf{Q}^{\textrm{H}}$ as
$$
\tilde{\mathbf{y}} = \mathbf{Q}^{\textrm{H}} \mathbf{y} = 
\mathbf{R} \mathbf{x} + \mathbf{Q}^{\textrm{H}} \mathbf{n} = \mathbf{R} \mathbf{x} + \hat{\mathbf{n}}.
$$

Assuming that $\mathbf{n}$ is i.i.d. Gaussian, then $\hat{\mathbf{n}}$ is i.i.d. Gaussian as well due to the orthogonal columns of $\mathbf{Q}$. Using that $\mathbf{R}$ is upper-triangular and that Algorithm \ref{alg:zf-sic} uses only the last element of $\hat{\mathbf{y}}$, we have the equation
$$
\hat{y}_{N_t} = r_{N_t, N_t} x_{N_t} + \hat{n}_{N_t}.
$$

This leads to the closed-form expression of the $K$ LLR values corresponding to the $N_t$-th symbol as
$$
\Lambda_{k, N_t} = \log \frac{\sum_{x_j, b=1} \exp -|\hat{y}_{N_t} - r_{N_t, N_t} x_j|^2}
        {\sum_{x_j, b=0} \exp -|\hat{y}_{N_t} - r_{N_t, N_t} x_j|^2}.
$$

Then, the previous equation also serves as a function $g(\hat{y}_{N_t}, r_{N_t, N_t}) = \Lambda_{:, N_t}$. Given that $\hat{y}_{N_t}$ is a complex scalar and $r_{N_t, N_t}$ is a real scalar, it follows that the LLR vector corresponding to the last spatial stream can be exactly represented by a three dimensional real vector. This proves the case $N_t = 1$. The equation for estimating the LLR values corresponding to the $K$-th symbol, given the previous $K-1$ estimates is then
$$
\hat{y}_{K} = r_{K, K} x_{K} + \sum_{k=1, \dots, K-1} r_{K, K-k} x_{K-k} + \hat{n}_{N_t}.
$$

Letting $\hat{t}_{K} = \hat{y}_{K} - \sum_{k=1, \dots, K-1} r_{K, K-k} x_{K-k}$, we have the compact equation
$$
\hat{t}_{K} = r_{K, K} x_{K} + \hat{n}_{N_t}.
$$

It follows that, given estimates for the previous $K-1$ symbols, the LLR values of the $K$-th symbol can be exactly represented by a vector with three real values, regardless of modulation order. Thus, by induction, $d_\textrm{ZF-SIC} = 3N_t$.

\end{appendices}

\end{document}